\documentclass[11pt]{article}

\makeatletter
\let\Otter@bibliographystyle\bibliographystyle
\renewcommand{\bibliographystyle}[1]{}
\makeatother

\usepackage{acl}

\makeatletter
\let\bibliographystyle\Otter@bibliographystyle

\def\@seccntformat#1{\csname the#1\endcsname.\quad}
\makeatother

\setcitestyle{numbers,square}

\usepackage{times}
\usepackage{latexsym}
\usepackage{svg}
\usepackage{stfloats}
\usepackage{float}
\usepackage{amsmath}
\usepackage{booktabs}
\usepackage{tabularx}
\usepackage{enumitem}

\usepackage{tikz}
\usetikzlibrary{shapes.geometric, arrows.meta, positioning, calc}

\usepackage[T1]{fontenc}
\usepackage[utf8]{inputenc}
\usepackage{microtype}
\usepackage{inconsolata}
\usepackage{graphicx}

\title{Otter: A Time-Aware, History-Conditioned Human Chess AI}

\author{Tarun Kumar S \\
  Peargent Labs \\
  Bangalore, India \\
  \texttt{tarun.peargentlabs@gmail.com} \\}

\renewenvironment{abstract}{
  \begin{center}
    \textbf{Abstract}
  \end{center}
  \itshape
}{}
\begin{document}
\maketitle
\begin{abstract}
\itshape

Forecasting human chess moves, not just engine-optimal ones, lies at the intersection of cognitive modeling and machine learning. Maia and Maia~2 showed that neural networks can replicate human playing styles~\cite{mcilroyyoung2020maia,tang2024maia2}, but Maia~2 treats each position independently, overlooking the sequential nature of games and the effect of time pressure on decision-making.

We present Otter, a 15.3M-parameter human chess AI that extends the Maia~2 paradigm with two additions: (1) a move history encoder that conditions predictions on the last 20 moves, capturing opening preferences, positional drift, and intra-game behavioral tendencies; and (2) a time control module that modulates predictions based on clock pressure. Otter is trained on 6.1~billion positions from 117~million Lichess rapid games for over 30~days on a single T4~GPU.

Otter achieves 55.23\% top-1 and 90.95\% top-5 accuracy, surpassing Maia~2 with far fewer parameters. Across 11~Elo brackets $(<1100 \text{ to } \geq 2000)$, accuracy peaks at 57.38\% in the 1900--1999 interval. Treating chess as a time-aware, sequential activity yields predictions closer to human play than position-only baselines, all with a smaller model. Our code, trained models, and complete training logs are publicly released.

\end{abstract}

\section{Introduction}

Chess engines like Stockfish~\cite{stockfish} and~\cite{lc0} find objectively optimal moves far beyond human capability. However, optimal play and human play differ fundamentally. Predicting the moves people actually make, accounting for habit, momentum, fatigue, and pressure, is a more impactful problem, with applications in personalized coaching, human-like AI opponents, cheat detection, and decision-making research under uncertainty.

Neural networks have proven effective at imitating human chess behavior. Maia introduced Elo-specific models trained on human games~\cite{mcilroyyoung2020maia}, outperforming weakened engines at predicting human moves. Maia~2 unified this into a single model with skill-coherent conditioning~\cite{tang2024maia2}, improving accuracy by nearly two percentage points. Both established that human move prediction is fundamentally a behavioral modeling problem.

Yet Maia~2 treats every position as an independent event, assuming the current board is a sufficient statistic for the next move. This Markov assumption~\cite{littman1994markov} is mathematically convenient but behaviorally wrong. Players carry momentum: a player who just blundered thinks differently from one slowly building an advantage. A player deep in prepared theory reacts differently from one in unfamiliar territory. Momentum, in-game tendencies, and positional drift all influence decisions, and a position-only model is blind to all of them~\cite{mcilroyyoung2021stylometry}.

Time is the second missing dimension. Under clock pressure, blunder rates spike, players favor forcing tactical continuations over complex positional plans, and decision-making transforms fundamentally~\cite{sunde2022speed,carow2025risk}. Yet existing models treat every position as if played with infinite time.

We present Otter,\footnote{Code and models: \url{https://github.com/PeargentLabs/otter-chess}.\\Training report: \href{https://api.wandb.ai/links/peargent-ai-labs/3mu4f1jv}{wandb.ai/peargent-ai-labs/Otter}.} a human chess AI that models play as an evolving, time-sensitive process. Otter extends skill conditioning with two components: a move history encoder (processing the last 20 moves via a Transformer encoder and cross-attention) and a time control module (encoding game format and remaining clock time) to jointly modulate predictions.

Both components are combined into a single conditioning signal accessible to every layer, enabling the entire model to depend on the full behavioral context. Despite having only 15.3M~parameters, Otter achieves state-of-the-art human move prediction accuracy, not through scale, but through modeling the human context behind each move.

\noindent \textbf{Our key contributions are:}
\begin{itemize}[noitemsep, topsep=0pt]

    \item A move history encoder using a Transformer to condition move predictions on the last 20 moves via cross-attention with board features, fused with skill and time conditioning, yielding a +5.24pp top-1 accuracy gain over a position-only baseline and showing that the Markov assumption is the primary bottleneck in position-only human chess models.
    \item A time control module combining game format and clock pressure as conditioning signals, adding +2.38pp beyond history alone. This improvement holds across all 11 Elo brackets, including the lowest where clock management is least deliberate, indicating that time pressure broadly influences human move selection.
    \item An ablation study showing both contributions are additive and consistent across all 11 Elo ranges, with a total improvement of +7.62pp over the position-only baseline, achieved without requiring a larger model.

\end{itemize}

\section{Related Work}
\textbf{Human Move Prediction.} Early human move prediction relied on modifying superhuman chess engines to play at a weaker level. This approach failed because engine-based weakening changes only objective playing strength without aligning with the human training distribution, meaning engine-generated moves do not predict human choices well. McIlroy-Young et al.\ overcame this by developing Maia~\cite{mcilroyyoung2020maia}, training nine separate models via supervised learning on human games across specific Elo brackets (1100--1900) based on the AlphaZero~\cite{silver2018alphazero} architecture. Tang et al.\ unified this in Maia~2~\cite{tang2024maia2} using categorical skill embeddings for both players combined through skill-aware attention, proving that a single model can continuously adapt its style across the entire Elo spectrum. However, Maia~2 treats every position as an independent event. It assumes the current board is a sufficient statistic for prediction, lacking any mechanism to represent how decision quality, game momentum, or clock pressure changes throughout a game. Subsequent work demonstrated that individual decision-making styles are detectable from move sequences~\cite{mcilroyyoung2021stylometry} and that per-player behavior models can be learned from game data~\cite{mcilroyyoung2022individual}, motivating the use of move history as a behavioral signal.\\
\textbf{Time Pressure in Human Chess.} Empirical studies show that time pressure degrades human decision-making, spikes blunder rates, and shifts players toward forcing tactical continuations over complex positional plans. Sunde et al.~\cite{sunde2022speed} analyze move-by-move quality and decision times across 80,000+ positions from 1,600 games, demonstrating that faster decisions yield poorer performance, consistent with sequential information acquisition under clock depletion. Carow and Witzig~\cite{carow2025risk} show that professional players under temporal stress favor risk-averse moves, while showing strategic loss aversion (greater risk-taking) when playing from a disadvantageous position. Leong et al.~\cite{leong2024time} show that experts adapt to temporal constraints using chunk-memory activation~\cite{chase_simon1973}, displaying distinct brain functional connectivity under stress. These cognitive dynamics vary across time controls (e.g., blitz pattern recognition vs.\ classical deep calculation) and shift dynamically within a single game as the clock runs down, making remaining clock time a critical predictor of human error.\\
\textbf{Sequence Modeling in Games.} Superhuman game-playing AIs use sequence information to track state and optimize play~\cite{crazyhouse2020}. AlphaZero represents board state as spatial planes repeated over an 8-step history to identify repetitions and en passant legality~\cite{silver2018alphazero}, a design also used by Chessformer~\cite{monroe2024chessformer}. Sequence-based models have also applied language modeling directly to move histories: Toshniwal et al.~\cite{toshniwal2022chess} track pieces via transformer sequences and show that full history attention is critical for predicting legal moves. Additionally, Ruoss et al.~\cite{ruoss2024amortized} show transformers can plan without explicit search, and Zhong et al.~\cite{huang2025ngram} use skill-specific n-grams to prove preceding move sequences carry predictive signal. However, these models use history to estimate objective board state or predict moves as a purely static context signal, failing to capture how a player's decision trajectory (such as the rise and fall in quality, momentum, and clock usage) affects their next move.
\begin{figure*}[!t]
    \centering
    \includegraphics[width=\textwidth]{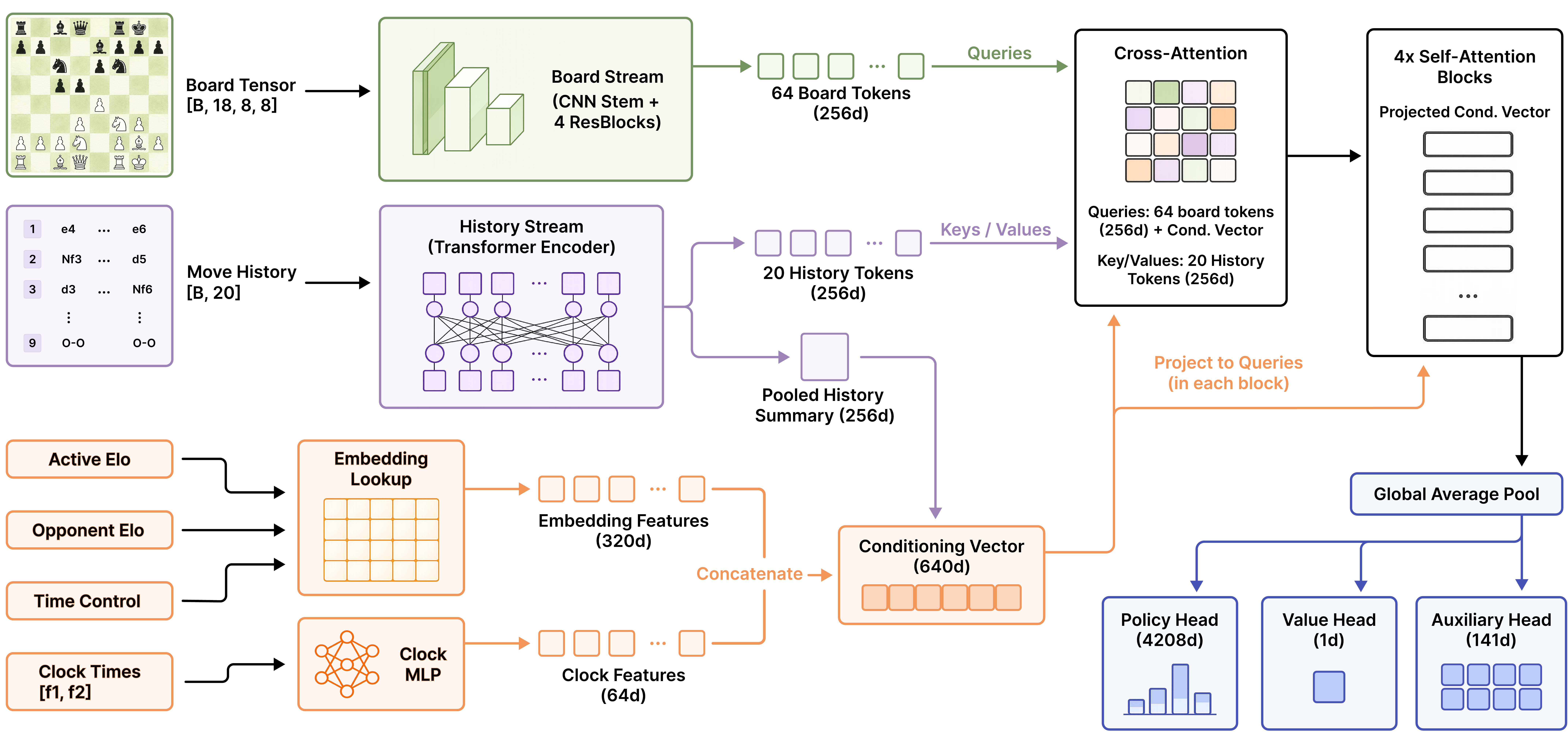}
    \caption{Overview of the Otter design. Our complete architecture encompasses four main stages: dual-stream encoding of the board state via a CNN and move sequences via a transformer; conditioning vector assembly by integrating Elo ratings, time controls, and clock characteristics with the pooled history; attention-based fusion of board and history representations with the conditioning vector projected directly into the queries; and joint multi-task prediction across policy, value, and auxiliary heads.}

    \label{fig:architecture}
\end{figure*}

\section{Architecture}

Otter is a 15.3~million-parameter neural network which simultaneously predicts move choice, game outcome, and move metadata from a chess position. The model takes six inputs, passes them through two parallel encoding streams, creates a unified conditioning vector, combines everything with a cross-attention block and four self-attention blocks, and generates three outputs from a shared pooled representation. Figure~\ref{fig:architecture} provides an overview of the complete pipeline.

\begin{table*}[!b]
\centering
\small
\caption{\textbf{Input summary of Otter.} The table outlines the six inputs processed by the model, including their shapes, dimensions, and semantic definitions.}
\label{tab:input_summary}
\begin{tabularx}{\textwidth}{l l l X}
  \toprule
  \textbf{Input Feature} & \textbf{Shape} & \textbf{Type / Range} & \textbf{Description} \\ \midrule
  Board Tensor & $[B, 18, 8, 8]$ & Binary (\texttt{float32}) & 18 planes encoding piece positions, castling rights, en passant, and active POV. \\
  Move History & $[B, 20]$ & Integer (\texttt{int64} in $[0, 4208]$) & Sequence of the last $K=20$ moves as vocabulary indices (0 for padding). \\
  Active Elo & $[B]$ & Integer (\texttt{int64} in $[0, 10]$) & Bucketed Elo rating of the active player ($<1100$ to $\ge2000$). \\
  Opponent Elo & $[B]$ & Integer (\texttt{int64} in $[0, 10]$) & Bucketed Elo rating of the opponent ($<1100$ to $\ge2000$). \\
  Time Control & $[B]$ & Integer (\texttt{int64} in $[0, 4]$) & Bucketed game time control format (bullet, blitz, rapid, classical, other). \\
  Clock Times & $[B, 2]$ & Float (\texttt{float32} in $[0, 1]$) & Remaining clock fractions for both the active player and the opponent. \\ \bottomrule
\end{tabularx}
\end{table*}

\subsection{Inputs}

The model receives six inputs per position, summarized in Table~\ref{tab:input_summary}: board tensor of shape [B, 18, 8, 8], a sequence of the last K=20 moves as vocabulary indices, a boolean padding mask, the active player's Elo bucket, the opponent's Elo bucket, a time control bucket, and two clock features representing remaining clock fractions. The board tensor contains 18~binary channels, detailed in Table~\ref{tab:channel_layout}: 12~for piece occupancy (6~piece types $\times$ 2~colors, always from the active player's perspective with the board mirrored for black), 4~for castling rights, 1~for the en-passant target square, and 1~indicating the active player's color. The vocabulary of moves consists of 4,208 legal UCI representations plus a padding token at index~0. \\
\indent Elo ratings are bucketed into 11~categories: bucket~0 covers <1100, buckets 1--9 cover 100-point intervals from 1100 to 2000, and bucket~10 covers $\geq 2000$. Time control is bucketed into 5~standard rapid formats. All moves in the history sequence are represented from the active player's perspective: the board is mirrored vertically and move squares flipped, ensuring consistent representation regardless of color.

\begin{table*}[t]
\centering
\small
\caption{\textbf{Board tensor channel layout.} Detailed breakdown of the 18 binary channels comprising the input board representation. All piece positions are encoded from the active player's canonical perspective.}
\label{tab:channel_layout}
\begin{tabularx}{\textwidth}{l l X}
  \toprule
  \textbf{Channel Index} & \textbf{Description} & \textbf{Value Details} \\ \midrule
  Channels 0--5 & Active Pieces & Binary planes: Pawn (0), Knight (1), Bishop (2), Rook (3), Queen (4), King (5). \\
  Channels 6--11 & Opponent Pieces & Binary planes: Pawn (6), Knight (7), Bishop (8), Rook (9), Queen (10), King (11). \\
  Channel 12 & Active K-Castling & All-ones plane if active player has kingside castling rights, all-zeros otherwise. \\
  Channel 13 & Active Q-Castling & All-ones plane if active player has queenside castling rights, all-zeros otherwise. \\
  Channel 14 & Opponent K-Castling & All-ones plane if opponent has kingside castling rights, all-zeros otherwise. \\
  Channel 15 & Opponent Q-Castling & All-ones plane if opponent has queenside castling rights, all-zeros otherwise. \\
  Channel 16 & En Passant Square & One-hot square plane with 1.0 at the en passant target square, all-zeros otherwise. \\
  Channel 17 & Side to Move & All-ones plane if original active player is White, all-zeros plane if Black. \\ \bottomrule
\end{tabularx}
\end{table*}

\subsection{Board Stream}

The board stream encodes the $18 \times 8 \times 8$ binary tensor into a sequence of 64~spatial tokens of dimension~256. The input passes through a CNN stem of three convolutional layers ($18 \to 64 \to 128 \to 256$ channels, $3 \times 3$ kernels, padding 1), each followed by batch normalization~\cite{batchnorm2015} and ReLU, with Dropout2d after the first two layers. Four residual blocks~\cite{resnet2016} follow, each containing two $256 \times 256$ convolutional layers with batch normalization, ReLU, and a skip connection.

The resulting $8 \times 8 \times 256$ feature map is flattened and transposed into 64~board tokens of dimension~256. Factored 2D positional embeddings are added: separate rank and file embedding tables (each $8 \times 256$) are maintained, and each token's positional embedding is the sum of its rank and file embeddings, giving the model explicit spatial awareness beyond what convolutions alone provide.

\subsection{History Stream}

Unlike AlphaZero's fixed 8-step positional look-back~\cite{silver2018alphazero}, this encoder is trained to detect behavioral patterns from the move sequence itself, capturing what each player is doing rather than only the board state.

\begin{figure}[!t]
    \centering
    \includegraphics[width=\columnwidth]{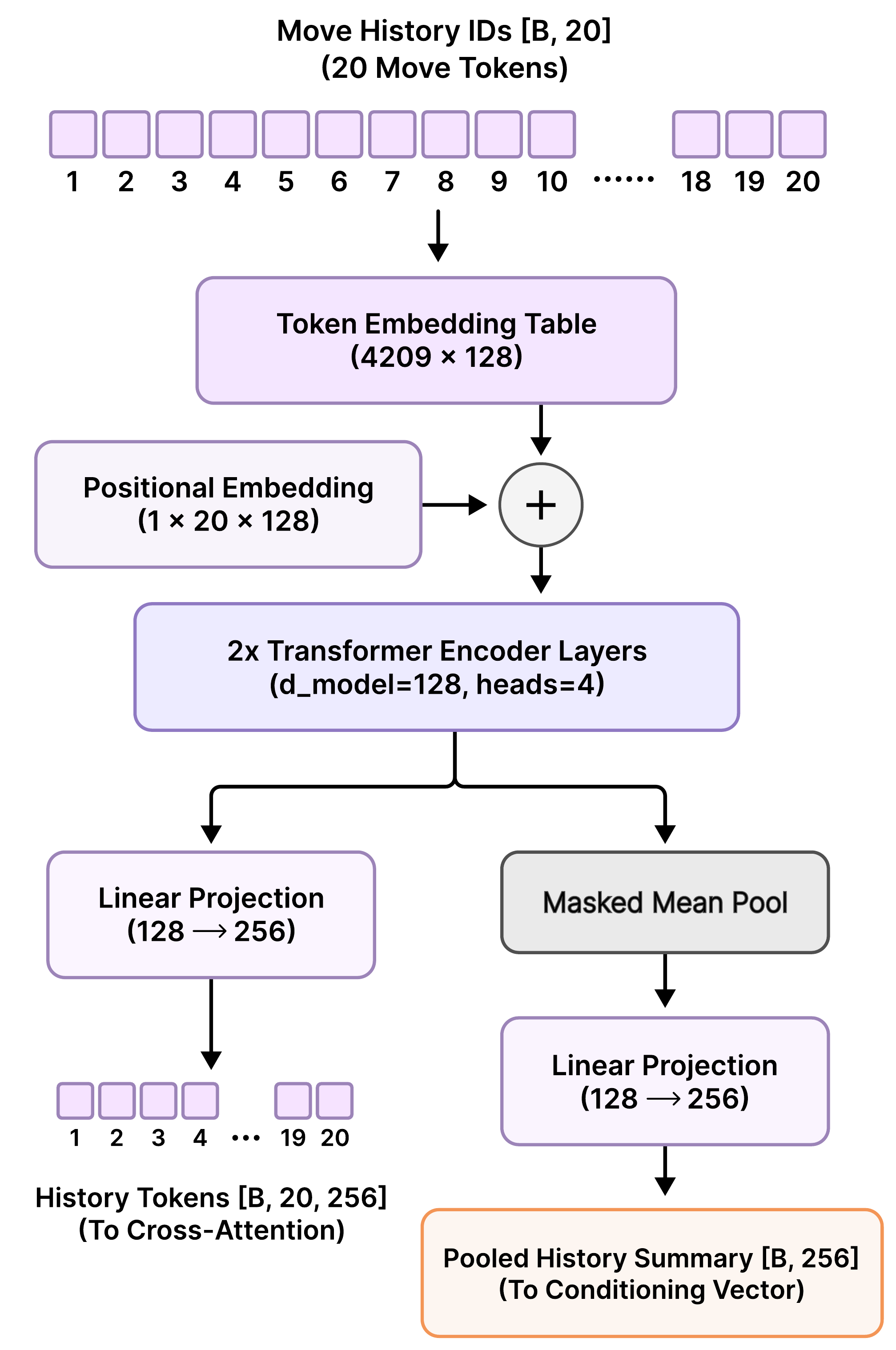}
  \caption{\textbf{Internal flow of the Move History Stream.} The move history is embedded, processed by a 2-layer Transformer, and split into token-level representations for cross-attention and a masked mean-pooled summary for the conditioning vector.}
    \label{fig:history_stream}
    \vspace{-10pt}
\end{figure}

The history stream, illustrated in Figure~\ref{fig:history_stream}, converts the last $K{=}20$ moves into two outputs: a token-level sequence for cross-attention with the board, and a single pooled summary vector for the conditioning signal.

\setlength{\textfloatsep}{6pt plus 2pt minus 2pt}
\setlength{\textfloatsep}{20pt plus 2pt minus 4pt}

Each move is mapped to an embedding via a lookup table of 4,209 entries (4,208 legal UCI moves plus one padding token at index~0, whose embedding is always zero). Learned positional embeddings are added, and the sequence is processed by a 2-layer Transformer encoder with 4~attention heads~\cite{attention2017} (dimension~32 each) and feedforward dimension~256. Padding positions are masked throughout.

The encoder output is mapped through two separate linear projections. The first projects each token from 128 to 256~dimensions, producing token-level representations for cross-attention. The second performs masked mean pooling over non-padding positions and projects the pooled 128~dimensions to 256, yielding a single summary vector of the game so far. This dual projection allows history to influence the model at two granularities: globally through the conditioning vector ("how has this game gone overall?") and locally through cross-attention, where individual board squares attend to specific past moves.

\subsection{Conditioning Vector and Clock Pressure Encoding}

A 640-dimensional conditioning vector is assembled from five components: active player Elo embedding (128-d, from \texttt{Embedding(11, 128)}), opponent Elo embedding (128-d), time control embedding (64-d, from \texttt{Embedding(5, 64)}), a clock feature vector (64-d), and the pooled history summary (256-d).

The time control bucket and clock features serve distinct roles: the bucket identifies the game format (structural constraints), while clock features capture the pressure at this specific moment. Together they provide temporal context at both the game and move level.

\textbf{Clock Pressure Encoding.} Human move selection depends not only on the time control format but also on the remaining clock time at the moment of decision. Otter represents the clock state as two normalized scalars:

\begin{equation}
f_1 = \frac{t_{\mathrm{remaining}}}{t_{\mathrm{base}}}
\qquad
f_2 = \frac{t_{\mathrm{increment}}}{t_{\mathrm{base}}}
\label{eq:time_features}
\end{equation}

Dividing by the base time standardizes time pressure across formats: a player with half their clock remaining ($f_1 = 0.5$) experiences comparable relative pressure whether playing a 10-minute or 15-minute game. The increment fraction $f_2$ captures structural time relief: a player with 30~seconds remaining ($f_1 \approx 0.05$) and no increment ($f_2 = 0$) faces immediate time hazard, whereas the same player with a 5-second increment ($f_2 \approx 0.008$) has a guaranteed baseline of thinking time per move. These two scalars are passed into a two-layer MLP:

\begin{equation}
\mathrm{clock} = \mathrm{MLP}([f_1, f_2])
\label{eq:clock}
\end{equation}
where $\mathrm{MLP}$ is a feedforward network structured as $\mathrm{Linear}(2 \to 64) \to \mathrm{ReLU} \to \mathrm{Linear}(64 \to 64) \to \mathrm{ReLU}$, producing a 64-dimensional clock representation concatenated into the conditioning vector alongside the Elo and history embeddings.

The complete conditioning vector is passed identically to every attention block in the fusion stage, simultaneously encoding player identity, opponent identity, time format, clock pressure, and game trajectory, making it the primary source of behavioral context throughout the model.

\subsection{Fusion}

The fusion stage integrates board and history tokens under the unified conditioning signal through a cross-attention block followed by four self-attention blocks.

\begin{figure}[!ht]
    \centering
    \includegraphics[width=\columnwidth]{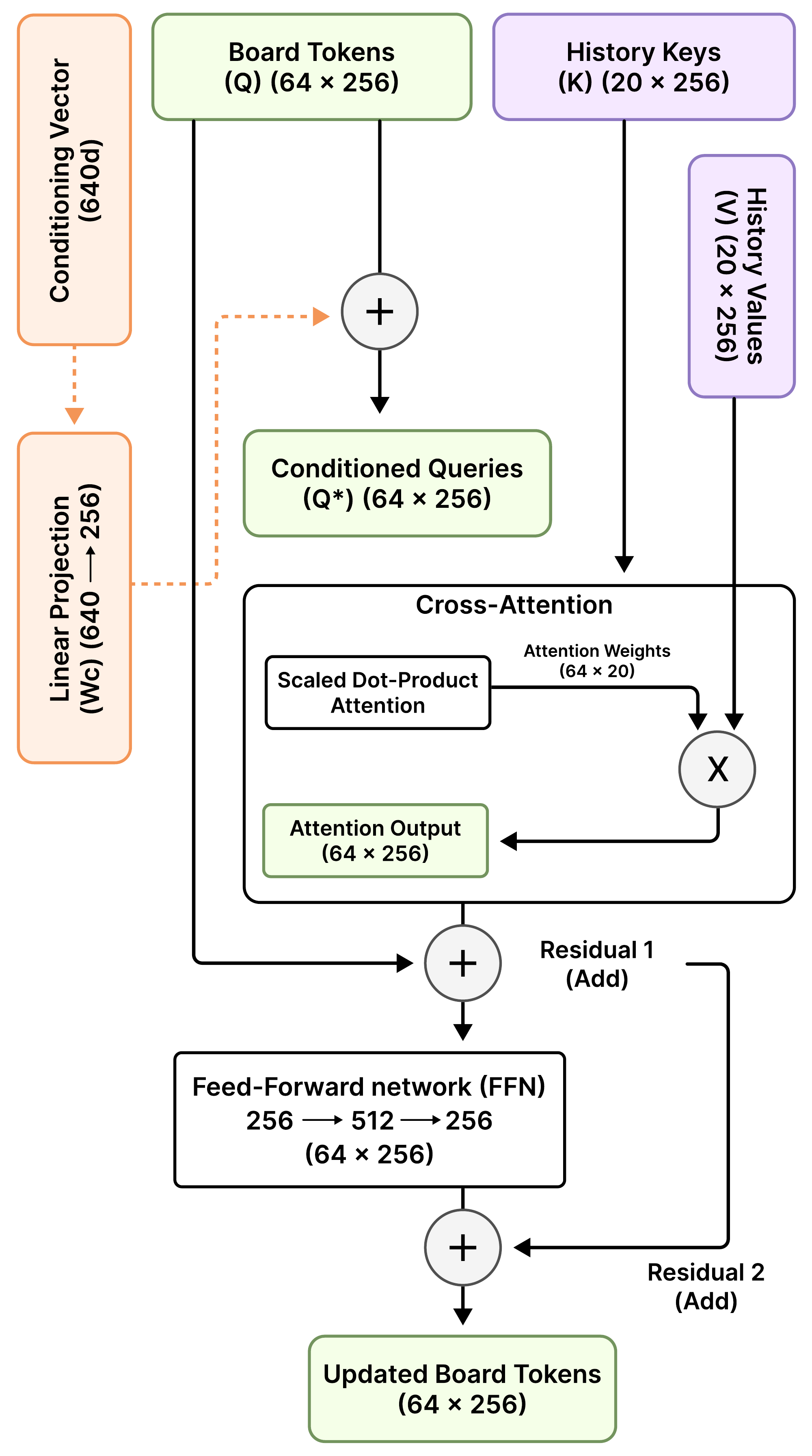}
  \caption{\textbf{Conditioned attention mechanism.} The 640-dimensional conditioning vector is projected via a learned linear layer $W_C$ and added to the board token queries to produce conditioned queries $Q^* = Q + \text{cond} \cdot W_C$. The conditioned queries attend to history tokens via cross-attention, followed by a feedforward network, with residual connections and pre-layer normalization throughout. The same mechanism is reused in the four subsequent self-attention blocks.}
    \label{fig:condition_mechanism}
\end{figure}

\textbf{Design Rationale: Query-Based vs.\ Normalization-Based Conditioning.} In vision and diffusion models, adaptive layer normalization (AdaLN-Zero)~\cite{peebles2023dit} is standard for conditioning. However, in game-playing agents like Maia~2~\cite{tang2024maia2}, injecting conditioning directly into attention queries has proven more effective. Queries dictate which board features the attention mechanism prioritizes, so injecting context there enables adaptive focus based on player style. We extend this to dynamic, multi-modal game context: Otter projects a composite conditioning vector (skill, time control, clock pressure, move history) into the query space of each attention head, enabling the attention mechanism to dynamically re-weight board features based on who is playing and the immediate temporal pressure.

In the cross-attention block, the 64~board tokens attend to the $K{=}20$ history tokens. Board tokens produce the queries; history tokens produce the keys and values. The conditioning vector is folded into the queries via a learned linear projection, as depicted in Figure~\ref{fig:condition_mechanism}:

\begin{equation}
    Q^* = Q + \text{cond} \cdot W_C
\end{equation}

where $Q$ is the board-token query matrix, $\text{cond}$ is the conditioning vector, $W_C$ is the learned projection into query space, and $Q^*$ is the conditioned query.

This allows the model to ask, for every board square, which aspects of game history are most relevant, with player identity and time pressure shaping the answer. A feedforward layer $(256 \rightarrow 512 \rightarrow 256)$ with GELU activation follows~\cite{gelu2016}, with residual connections and pre-layer normalization throughout.

Four self-attention blocks continue with the same conditioned-query technique: the conditioning vector is projected and added to queries in each block, so the model refines spatial relations among all 64 squares while remaining aware of skill, time pressure, and game history. Each block uses 8~heads of size~32 and a feedforward network $(256 \rightarrow 1024 \rightarrow 256)$ with GELU and residual connections. After these blocks, global average pooling merges the 64~tokens into a single 256-dimensional vector for the output heads. Crucially, the conditioning vector is injected into every attention block (cross-attention plus all four self-attention blocks), ensuring time awareness impacts spatial reasoning at every stage.

\subsection{Output Heads}

Three heads independently map the 256-dimensional pooled representation. The \textbf{policy head} outputs a distribution over 4,208 legal moves via a two-layer MLP $(256 \rightarrow 1024 \rightarrow 4{,}208)$ with ReLU and dropout~\cite{dropout2014}; illegal moves are masked to $-\infty$ before softmax. The \textbf{value head} estimates the game result via a mini-MLP $(256 \rightarrow 64 \rightarrow 1)$ with Tanh output, yielding a scalar in $[-1, +1]$. The \textbf{auxiliary head} predicts 141~binary move properties (moving piece type (6), captured piece type (6), check flag (1), from-square (64), to-square (64)) via a two-layer MLP $(256 \rightarrow 512 \rightarrow 141)$ with Sigmoid activation. Table~\ref{tab:architecture} summarizes the parameter allocation across all components.

\begin{table}[H]
\centering
\caption{\textbf{Architecture summary} (15.3M parameters).}
\label{tab:architecture}
\resizebox{\columnwidth}{!}{%
\begin{tabular}{@{}llr@{}}
\toprule
\textbf{Component} & \textbf{Details} & \textbf{Params} \\
\midrule
\multicolumn{3}{@{}l}{\textit{Board Stream}} \\
\quad CNN Stem & $18{\to}64{\to}128{\to}256$ & \\
\quad Res.\ Blocks & $\times$4, $256{\to}256$ & \\
\quad Pos.\ Embed & Rank + File & 5.11M \\
\midrule
\multicolumn{3}{@{}l}{\textit{History Stream}} \\
\quad Embedding & $4209 \times 128$ & \\
\quad Transformer & $\times$2, 4 heads & \\
\quad Projections & $128{\to}256$ each & 0.87M \\
\midrule
\multicolumn{3}{@{}l}{\textit{Conditioning (640d)}} \\
\quad Elo & Emb(11, 128) $\times$ 2 & \\
\quad Time / Clock & Emb(5, 64) + Clock MLP & 0.01M \\
\midrule
\multicolumn{3}{@{}l}{\textit{Fusion}} \\
\quad Cross-Attn & $\times$1, 8 heads & 0.69M \\
\quad Self-Attn & $\times$4, 8 heads & 3.81M \\
\midrule
\multicolumn{3}{@{}l}{\textit{Output Heads}} \\
\quad Policy & $256{\to}1024{\to}4208$ & 4.58M \\
\quad Value & $256{\to}64{\to}1$ & 0.02M \\
\quad Auxiliary & $256{\to}512{\to}141$ & 0.20M \\
\midrule
\textbf{Total} & & \textbf{15.3M} \\
\bottomrule
\end{tabular}%
}
\end{table}

\section{Training}

We describe the training methodology, preprocessing pipeline, and optimization schedule used to train Otter's parameters. The model is trained via supervised learning on 117~million Lichess rapid games (6.1~billion positions), using mixed precision on a single T4~GPU to jointly predict players' next moves, expected game outcome, and auxiliary move properties. Complete training settings and data pipelines are described in the following subsections.

\subsection{Dataset}
Otter is trained exclusively on rated rapid games from the Lichess 2024 public database~\cite{lichess_db}. Lichess provides complete game records in PGN format with move-level clock times, time controls, and player ratings. Per Lichess's formula $(\text{base seconds} + 40 \times \text{increment})$, this corresponds to estimated game durations of 8--25 minutes. The dataset is predominantly 10+0 games (${\sim}75\%$), followed by 10+5 and 15+10. Bullet and blitz games are excluded because their extreme time pressure produces fundamentally different move selection behavior; classical games are excluded due to their scarcity on Lichess. The final dataset comprises 117~million games and approximately 6.1~billion positions; Table~\ref{tab:dataset_stats} summarizes the training, validation, and test splits. Games with missing clock data, incomplete move records, or players outside the supported Elo range are filtered out. We report the rating-bracket matchup distribution of the final training dataset in Table~\ref{tab:elo_matchup}. Because we flip Black's moves to play as the active player, the matchup statistics between any two rating brackets are symmetric; we therefore combine these symmetric counts and suppress the redundant upper half of the table. The diagonal dominance of the table reflects Lichess's rating-balanced pairing system.

\begin{table*}[b]
\centering
\caption{\textbf{Dataset statistics.} Summary of the training, validation, and test datasets extracted from Lichess.}
\label{tab:dataset_stats}
\begin{tabular}{@{}llllrr@{}}
\toprule
\textbf{Split} & \textbf{Source} & \textbf{Period} & \textbf{Format} & \textbf{Total Games} & \textbf{Total Positions} \\
\midrule
Training & Lichess & 2024 & Rapid ($8{+}0$ to $15{+}10$) & 117M & $\sim$6.1B \\
Validation & Lichess & Jan 2025 & Rapid ($8{+}0$ to $15{+}10$) & ---$^{\text{a}}$ & 204,800 \\
Test (Ablation / Exp) & Lichess & Feb 2025 & Rapid ($8{+}0$ to $15{+}10$) & ---$^{\text{b}}$ & 1,100,000 \\
\bottomrule
\end{tabular}
\smallskip
\flushleft
\footnotesize
$^{\text{a}}$ Streamed deterministically with a fixed seed during training (100 steps of batch size 2,048).\\
$^{\text{b}}$ The test set consists of a fixed, balanced subset of positions sampled from February 2025 games.
\end{table*}

\begin{table*}[b]
\centering
\caption{\textbf{2024 training set Elo matchup distribution.} Number of games per rating-pair bracket. The upper triangle is suppressed by symmetry because Black's moves are flipped to play as the active player (White). Total training games: 117,235,902.}
\label{tab:elo_matchup}
\resizebox{\textwidth}{!}{%
\begin{tabular}{@{}lrrrrrrrrrrr@{}}
\toprule
\textbf{Black} & \multicolumn{11}{c}{\textbf{White Elo Bracket}} \\
\cmidrule(l){2-12}
 & \textbf{$<$1100} & \textbf{1100--1199} & \textbf{1200--1299} & \textbf{1300--1399} & \textbf{1400--1499} & \textbf{1500--1599} & \textbf{1600--1699} & \textbf{1700--1799} & \textbf{1800--1899} & \textbf{1900--1999} & \textbf{$\geq$2000} \\
\midrule
\textbf{$<$1100}      & 14,284,462 & ---        & ---        & ---        & ---        & ---        & ---        & ---        & ---        & ---        & ---        \\
\textbf{1100--1199}   &  3,069,295 &  4,186,924 & ---        & ---        & ---        & ---        & ---        & ---        & ---        & ---        & ---        \\
\textbf{1200--1299}   &    268,631 &  3,149,303 &  5,318,190 & ---        & ---        & ---        & ---        & ---        & ---        & ---        & ---        \\
\textbf{1300--1399}   &     85,625 &    213,561 &  3,503,137 &  6,334,001 & ---        & ---        & ---        & ---        & ---        & ---        & ---        \\
\textbf{1400--1499}   &     52,366 &     55,034 &    212,655 &  3,689,289 &  7,038,315 & ---        & ---        & ---        & ---        & ---        & ---        \\
\textbf{1500--1599}   &     59,921 &     73,672 &    160,208 &    408,664 &  4,121,176 &  7,887,600 & ---        & ---        & ---        & ---        & ---        \\
\textbf{1600--1699}   &     21,161 &     23,112 &     44,843 &     84,696 &    237,567 &  4,138,328 &  7,711,141 & ---        & ---        & ---        & ---        \\
\textbf{1700--1799}   &     17,422 &      9,650 &     26,426 &     47,391 &     82,293 &    340,431 &  4,002,956 &  7,192,832 & ---        & ---        & ---        \\
\textbf{1800--1899}   &     12,571 &      9,288 &      8,053 &     19,728 &     40,398 &     99,160 &    257,911 &  3,748,788 &  5,994,453 & ---        & ---        \\
\textbf{1900--1999}   &     13,412 &      7,107 &      7,231 &      9,620 &     21,455 &     49,979 &     64,228 &    277,227 &  3,233,845 &  4,348,356 & ---        \\
\textbf{$\geq$2000}   &     24,227 &     11,688 &     12,348 &     15,656 &     22,554 &     47,074 &     56,123 &     87,873 &    328,711 &  2,719,330 &  7,535,230 \\
\bottomrule
\end{tabular}%
}
\end{table*}

\subsection{Preprocessing}
Each game is converted into per-position training examples. For every position, we extract: the 18-channel board tensor, the last $K{=}20$ moves as vocabulary indices, active and opponent Elo buckets, the time control category, and remaining clock fractions for both players. The board tensor is always constructed from the active player's perspective (mirrored for black). Move history is canonicalized in the same frame, with squares flipped for black moves. Sequences shorter than 20 moves are padded with the \texttt{PAD} token (index~0), accompanied by a boolean mask excluding padding from attention. Legal move masks are precomputed per position.

\subsection{Loss Function}
The three output heads are trained jointly:

\begin{equation}
\mathcal{L} = \mathcal{L}_{\text{policy}} + 0.25 \cdot \mathcal{L}_{\text{value}} + 0.5 \cdot \mathcal{L}_{\text{aux}}
\label{eq:loss}
\end{equation}

The policy loss $\mathcal{L}_{\text{policy}}$ is cross-entropy between the predicted move distribution and the actual human move. The value loss $\mathcal{L}_{\text{value}}$ is MSE between predicted and actual game outcomes ($+1$ win, $0$ draw, $-1$ loss). The auxiliary loss $\mathcal{L}_{\text{aux}}$ is binary cross-entropy on the 141~move-property predictions.

The policy head carries full weight ($1.0$) as the primary target. The value head ($0.25$) serves as a regularizer encouraging game-state awareness. The auxiliary head ($0.5$) provides structured supervision grounding the policy representation in concrete move semantics.

In practice, $\mathcal{L}_{\text{value}}$ does not decrease significantly. Predicting game outcomes from single mid-game positions in human rapid chess is inherently high-entropy due to blunders and time pressure. The value head is retained as an auxiliary regularizer but is not reported as a key metric. $\mathcal{L}_{\text{policy}}$ converges steadily, with residual variance reflecting natural stochasticity in human move choice rather than training failure.

\subsection{Optimizer and Schedule}

We optimize Otter using the AdamW optimizer~\cite{adamw2019} with a base learning rate of $10^{-4}$ and weight decay of $10^{-5}$. To stabilize the initial training phase and prevent early divergence in the Transformer's cross-attention layers, we employ a linear learning rate warmup over the first 10\% of the optimization budget (the first 300{,}000 steps). Following warmup, a cosine annealing schedule~\cite{sgdr2017} decays the learning rate to a minimum of $10^{-6}$ at 3{,}000{,}000 steps. Gradients are clipped to a maximum $L_2$ norm of 1.0. Mixed-precision training~\cite{mixedprecision2018} (PyTorch AMP with \texttt{GradScaler}~\cite{pytorch2019}) is used to accelerate throughput and manage memory on the NVIDIA T4~GPU.

\begin{table}[H]
\centering
\caption{\textbf{Training configuration.} Summary of hyperparameters, optimization settings, and compute resources.}
\label{tab:training_config}
\resizebox{\columnwidth}{!}{%
\begin{tabular}{@{}ll@{}}
\toprule
\textbf{Hyperparameter / Setting} & \textbf{Value} \\
\midrule
Optimizer & AdamW \\
Base Learning Rate & $1\times 10^{-4}$ \\
Final Learning Rate & $1\times 10^{-6}$ \\
Learning Rate Schedule & Cosine annealing ($10\%$ linear warmup) \\
Weight Decay & $1\times 10^{-5}$ \\
Gradient Clipping & Max $L_2$ norm $1.0$ \\
Batch Size & $2048$ \\
Training Steps & $3,000,000$ \\
Total Positions Seen & $\sim$6.1B \\
Precision & Mixed (FP16 AMP with GradScaler) \\
\midrule
Hardware & $1\times$ NVIDIA T4 GPU \\
Training Wall Time & $\sim$30 days \\
\bottomrule
\end{tabular}%
}
\end{table}
\subsection{Hardware and Training Time}

Training ran on a single NVIDIA T4~GPU for approximately 30~days: 3~million steps at batch size~2048, processing $\sim$6.1~billion positions total. No distributed training or gradient accumulation was used. The T4's 16~GB VRAM accommodated this batch size under mixed precision. The complete training configuration, optimization parameters, and hardware details are summarized in Table~\ref{tab:training_config}.

\section{Experiments and Results}

We present a series of empirical evaluations and key results to assess Otter's ability to model human chess decisions. We evaluate the model's accuracy across various skill levels, compare it directly to the state-of-the-art Maia 2 baseline, and conduct ablation studies to isolate the impact of history and temporal features. Additionally, we analyze the model's sensitivity to history window length and game phases. Our findings demonstrate that incorporating behavioral move history and time pressure context yields consistent, universal improvements in move prediction accuracy across all rating brackets, outperforming the position-only baseline while using a smaller parameter footprint.

\subsection{Evaluation Methodology}
All models are evaluated on data outside the 2024 training period to prevent temporal leakage. January 2025 Lichess rapid games serve as the validation set during training (full model: 55.57\% top-1 at convergence). Final evaluation uses an unseen test set of 1{,}100{,}000 positions from February 2025 Lichess rapid games, equally divided among 11~Elo brackets (100{,}000 per bracket).

\textbf{Top-1 accuracy} is the fraction of positions where the model's highest-probability prediction matches the human move. \textbf{Top-5 accuracy} is the fraction where the human move falls within the five highest-probability predictions. Illegal moves are always masked to $-\infty$ before softmax. All three ablation variants (base, history only, full) are evaluated on identical test positions. Maia~2 comparisons use publicly reported figures from~\cite{tang2024maia2}.

\subsection{Comparison Against Maia~2}

Otter is compared to Maia~2, the state-of-the-art position-only human chess model as of February~2026~\cite{tang2024maia2}, a unified skill-conditioned model trained on 9.1~billion positions from 169~million games. The comparison assesses how much behavioral context improves over the best existing position-only method.

Specifically, Maia~2 reports a macro-averaged move prediction accuracy of 53.25\%, achieving 51.72\% for Skilled players (up to 1600 Elo), 54.15\% for Advanced players (1600--2000 Elo), and 53.87\% for Master players (2000$+$ Elo). Otter achieves consistent improvements across all corresponding cohorts: 54.66\% ($+$2.94\,pp) for Skilled, 56.32\% ($+$2.17\,pp) for Advanced, and 57.09\% ($+$3.22\,pp) for Master players.

The Base model (position-only) achieves 47.61\%, which is 5.64\,pp below Maia~2's 53.25\%, confirming that architectural differences alone do not explain the improvement. As summarized in Table~\ref{tab:maia_comparison}, the full model at 55.23\% surpasses Maia~2 by $+$1.98\,pp despite using ${\sim}$31\% less training data and 34\% fewer parameters. This improvement is  entirely attributable to behavioral context, as the ablation study confirms. Note that this comparison is not fully controlled for training data distribution.

\begin{table}[H]
\centering
\caption{\textbf{Main comparison against Maia 2.} We compare parameter size, training database scale, and overall human move prediction accuracy.}
\label{tab:maia_comparison}
\resizebox{\columnwidth}{!}{%
\begin{tabular}{@{}lrrrr@{}}
\toprule
\textbf{Model} & \textbf{Params} & \textbf{Train Games} & \textbf{Top-1 Acc.} & \textbf{Top-5 Acc.} \\
\midrule
Maia 2\textsuperscript{a} & 23.3M & 169M & $53.25\%$ & --- \\
\textbf{Otter} (Ours) & \textbf{15.3M} & \textbf{117M} & $\mathbf{55.23\%}$ & $\mathbf{90.95\%}$ \\
\bottomrule
\end{tabular}%
}
\smallskip
\flushleft
\footnotesize
\textsuperscript{a} Maia 2 numbers are cited directly from Tang et al. \cite{tang2024maia2}.
\end{table}

\begin{table*}[b]
\centering
\caption{\textbf{Per-bracket move prediction accuracy.} Top-1 and top-5 accuracy for the Base, History-only, and Full model variants across all 11 Elo brackets. $\Delta$\,Hist and $\Delta$\,Full denote percentage point improvements over the Base model. The Overall row is the macro-average across all 1{,}100{,}000 test positions.}
\label{tab:per_bracket}
\resizebox{\textwidth}{!}{%
\begin{tabular}{@{}lrcccccrr@{}}
\toprule
\textbf{Elo Bracket} & \textbf{$n$} & \textbf{Base Top-1} & \textbf{Hist.\ Top-1} & \textbf{Full Top-1} & \textbf{$\Delta$ Hist} & \textbf{$\Delta$ Full} & \textbf{Base Top-5} & \textbf{Full Top-5} \\
\midrule
$<$1100      & 100{,}000   & 42.26\% & 47.42\% & 49.48\% & $+$5.16\,pp & $+$7.22\,pp & 82.01\% & 86.08\% \\
1100--1199   & 100{,}000   & 45.58\% & 50.98\% & 53.54\% & $+$5.40\,pp & $+$7.96\,pp & 85.45\% & 89.67\% \\
1200--1299   & 100{,}000   & 46.95\% & 52.24\% & 54.61\% & $+$5.29\,pp & $+$7.66\,pp & 86.41\% & 90.28\% \\
1300--1399   & 100{,}000   & 47.72\% & 52.85\% & 55.19\% & $+$5.13\,pp & $+$7.47\,pp & 87.07\% & 90.75\% \\
1400--1499   & 100{,}000   & 47.85\% & 52.87\% & 55.29\% & $+$5.02\,pp & $+$7.44\,pp & 87.14\% & 90.98\% \\
1500--1599   & 100{,}000   & 47.60\% & 52.97\% & 55.39\% & $+$5.37\,pp & $+$7.79\,pp & 87.36\% & 91.04\% \\
1600--1699   & 100{,}000   & 48.84\% & 54.05\% & 56.59\% & $+$5.21\,pp & $+$7.75\,pp & 88.06\% & 91.94\% \\
1700--1799   & 100{,}000   & 48.69\% & 53.90\% & 56.46\% & $+$5.21\,pp & $+$7.77\,pp & 88.50\% & 92.11\% \\
1800--1899   & 100{,}000   & 49.16\% & 54.48\% & 56.85\% & $+$5.32\,pp & $+$7.69\,pp & 88.75\% & 92.25\% \\
1900--1999   & 100{,}000   & 49.68\% & 54.94\% & 57.38\% & $+$5.26\,pp & $+$7.70\,pp & 89.36\% & 92.85\% \\
$\geq$2000   & 100{,}000   & 49.35\% & 54.61\% & 56.80\% & $+$5.26\,pp & $+$7.45\,pp & 89.01\% & 92.49\% \\
\midrule
\textbf{Overall} & \textbf{1{,}100{,}000} & \textbf{47.61\%} & \textbf{52.85\%} & \textbf{55.23\%} & $\mathbf{+}$\textbf{5.24\,pp} & $\mathbf{+}$\textbf{7.62\,pp} & \textbf{87.19\%} & \textbf{90.95\%} \\
\bottomrule
\end{tabular}%
}
\end{table*}
\subsection{Per-Bracket Analysis}

\begin{figure}[H]
  \centering
  \includegraphics[width=\columnwidth]{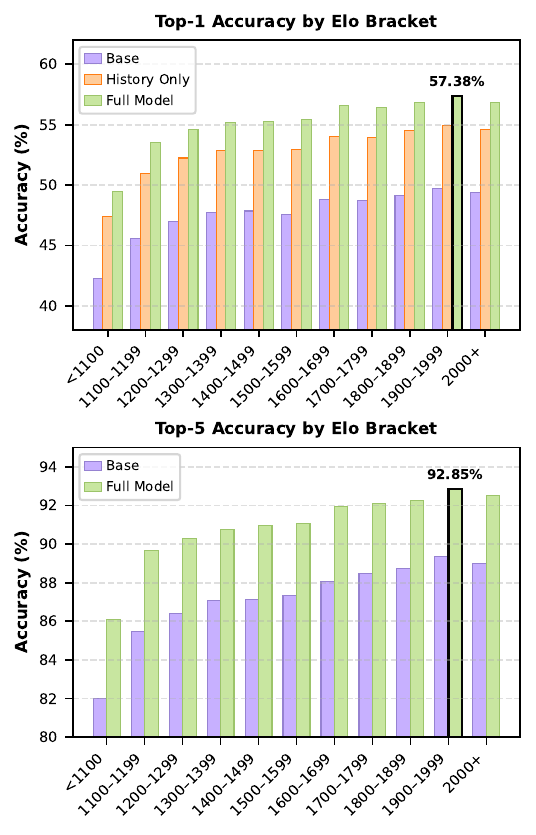}
  \caption{\textbf{Per-bracket accuracy across ablation variants.} Top-1 accuracy (top) for all three variants and top-5 accuracy (bottom) for Base and Full model across all 11 Elo brackets. Peak top-1 of 57.38\% and top-5 of 92.85\% are both achieved in the 1900--1999 bracket.}
  \label{fig:bracket_ablation}
\end{figure}

Otter is analyzed across all 11 Elo brackets to verify that accuracy improvements are uniformly distributed rather than concentrated at specific rating levels. Figure~\ref{fig:bracket_ablation} shows per-bracket top-1 and top-5 accuracy for all three variants; Table~\ref{tab:per_bracket} provides the full numerical breakdown.

Full model's top-1 accuracy correlates positively with the Elo, rising from 49.48\% ($<$1100) to 57.38\% (1900--1999), with a slight drop to 56.80\% ($\geq$2000). This pattern holds across all of the three 

\noindent variants and is consistent with stronger players making more contextually predictable moves given sufficient behavioral context.

The improvement from adding history and time is remarkably uniform across Elo, ranging from $+$7.22\,pp ($<$1100) to $+$7.96\,pp (1100--1199), with no bracket falling below $+$7\,pp. Top-5 accuracy ranges from 86.08\% ($<$1100) to 92.85\% (1900--1999), with the base-to-full deltas of $+$3.48\,pp to $+$4.22\,pp, confirming that the full model's probability distributions are better calibrated across all skill levels.

\subsection{Ablation Study}

To isolate the contribution of each component, three Otter variants are evaluated on the same 1,100,000-position test set:

\begin{itemize}
    \item \textbf{Base} (no history, no time): board tensor + skill conditioning only, equivalent to a position-only model with Otter's architecture.
    \item \textbf{History only}: base + move history encoder and cross-attention, no time module.
    \item \textbf{History and time} (full model): all components active.
\end{itemize}

The ablation reveals three key insights:

\textbf{Insight 1: The Markov assumption is the primary bottleneck \cite{littman1994markov}.} Adding move history alone yields $+$5.24\,pp overall, from 47.61\% to 52.85\%. The history-only model (52.85\%) approaches Maia-2's 53.25\% \cite{tang2024maia2} despite 34\% fewer parameters and less training data. The position-only assumption, not model capacity or data volume, is the main limitation.

\textbf{Insight 2: Time control is a separate, additive signal.} Adding the time module on top of history yields an extra $+$2.38\,pp (52.85\% $\to$ 55.23\%), observed across all brackets, from $+$2.06\,pp ($<$1100) to $+$2.54\,pp (1600--1699). That the effect holds even at the lowest Elo levels, where clock management is least deliberate, suggests time pressure broadly influences human move selection.

\textbf{Insight 3: Both contributions are additive and universal.} The combined $+$7.62\,pp improvement appears in all 11 brackets with no exceptions (minimum $+$7.22\,pp).  No bracket shows degradation

\noindent  from adding either component. This rules out gains  being attributable to particular game subsets and supports the view that sequential context and time pressure are fundamental features of human chess decision-making.

\subsection{Accuracy by Game Phase}
\begin{figure}[H]
    \centering
    \includegraphics[width=\columnwidth]{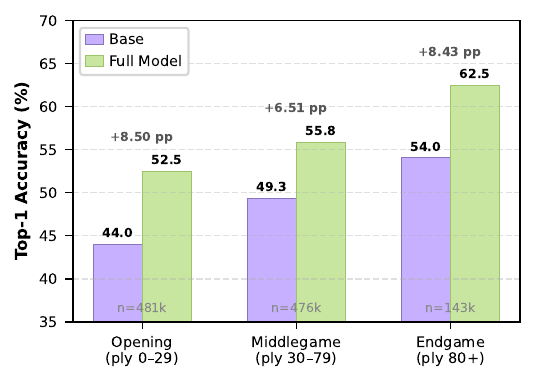}
    \caption{\textbf{Move prediction accuracy by game phase.} Comparison of the position-only Base model and the Full model across openings (ply 0--29), middlegames (ply 30--79), and endgames (ply 80+). Absolute percentage point improvements are annotated above each pair. The sample size $n$ represents the total number of test board positions evaluated in each respective phase.}
    \label{fig:game_phases}
\end{figure}

Positions are divided into three phases: opening (ply 0--29), middlegame (ply 30--79), and endgame (ply 80+). Figure~\ref{fig:game_phases} compares the Base and Full models across all three phases.

\begin{itemize}
    \item \textbf{Opening}: base 43.98\%, full 52.48\% ($+$8.50\,pp, $n{=}480{,}870$). Even a few moves of history reveal a player's repertoire and style preferences. Notably, this advantage is even more pronounced in the first five moves of the game (ply 0--9, $n{=}167{,}519$), where the full model outperforms the base model by $\mathbf{+12.63}$\,\textbf{pp} (52.09\% vs.\ 39.46\%, top-5: 90.74\% vs.\ 86.34\%). This demonstrates that the history encoder can identify style preferences and opening repertoires even from extremely short, heavily padded sequences.

    \item \textbf{Middlegame}: base 49.33\%, full 55.84\% ($+$6.51\,pp, $n{=}475{,}653$). Improvement is genuine but smaller due to positional diversity.

    \item \textbf{Endgame}: base 54.03\%, full 62.46\% ($+$8.43\,pp, $n{=}143{,}477$). Endgame positions are highly path-dependent, and the history encoder captures the chain of events the base model cannot see.
\end{itemize}

\subsection{History Window Size Sensitivity}
To justify $K{=}20$, the full model is evaluated with inference-time history truncation at $K{=}5$, $K{=}10$, and $K{=}20$, with all other components held constant. Oldest tokens are replaced with zeros to simulate shorter windows.

\begin{figure}[H]
    \centering
    \includegraphics[width=\columnwidth]{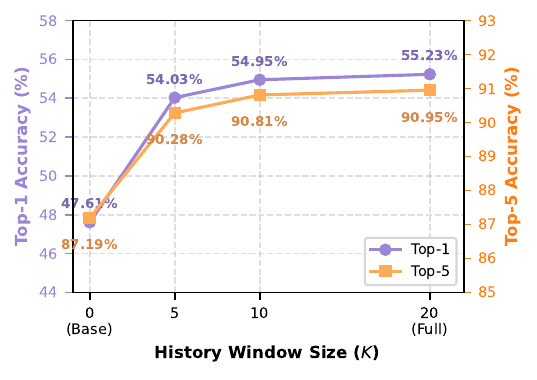}
    \caption{\textbf{History window size sensitivity.} Top-1 accuracy as a function of the history window size $K$. $K=0$ represents the no-history Base model baseline (47.61\%, plotted as a dashed line). Performance rises steeply at $K=5$ and flattens out towards the chosen configuration of $K=20$.}
    \label{fig:history_sensitivity}
\end{figure}

Results show diminishing returns (Figure~\ref{fig:history_sensitivity}): $K{=}5$ reaches 54.03\%, $K{=}10$ reaches 54.95\%, and $K{=}20$ reaches 55.23\%. Relative to the no-history baseline (47.61\%), $K{=}5$ already retains 84\% of the total history advantage ($+$6.42\,pp out of $+$7.62\,pp). The gain from $K{=}5$ to $K{=}10$ is $+$0.92\,pp, and from $K{=}10$ to $K{=}20$ only $+$0.28\,pp, showing clear saturation at $K{=}20$.

\begin{figure*}[b]
    \centering
    \includegraphics[width=\textwidth]{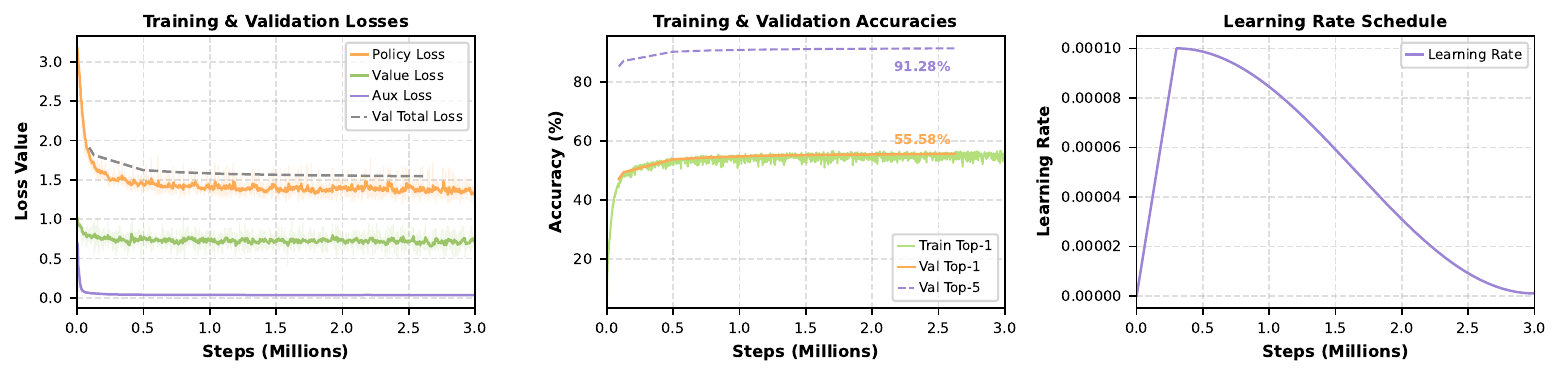}
    \caption{\textbf{Training dynamics and convergence curves across 3.0M optimization steps.} Panel 1 plots the policy, auxiliary, and value training losses (solid lines) alongside the validation total loss (dashed line). Panel 2 compares the training top-1 move prediction accuracy with validation top-1 (solid) and top-5 (dashed) accuracies evaluated on the January 2025 validation set, with final values annotated. Panel 3 shows the learning rate schedule, detailing the initial 10\% linear warmup (steps 0--300k) followed by cosine annealing decay to $10^{-6}$. The validation metrics closely track training metrics throughout the run, confirming stable regularization and the absence of overfitting.}
    \label{fig:training_dynamics}
\end{figure*}

\subsection{Training Dynamics}
We analyze the training progression of Otter over the 3.0M steps of optimization, shown in Figure~\ref{fig:training_dynamics}.

The auxiliary loss converges smoothly, confirming the model learns move-property prediction as a stable secondary task. 
Policy loss converges steadily with expected noise; human move choice is inherently stochastic, imposing a floor on achievable loss, and the observed variance is normal. The value loss remains flat and noisy, which is unsurprising: predicting game outcomes from single positions in human rapid chess is high-entropy due to blunders and time pressure. The value head is retained as a regularizer but not reported as a metric.

Validation accuracy (January 2025, stabilizing at 55.57\% top-1) tracks training accuracy closely throughout, with no divergence even after 30 days, confirming no overfitting. All reported evaluation numbers are from the February 2025 test set.

\subsection{Error Analysis}

Otter's errors are consistent with the nature of human chess behavior, revealing inherent limits of behavioral context modeling.

When the top-1 prediction misses, the predicted move is typically a reasonable alternative, not an arbitrary choice. The large top-1/top-5 gap in lower Elo brackets reflects less consolidated opening knowledge and more variable move choice. The 86.08\% top-5 accuracy in the $<$1100 bracket shows the model correctly identifies the plausible move set even when the specific choice is unpredictable.

Blunders are systematically underpredicted. Otter is trained to predict modal human behavior at each Elo level, so it favors sensible moves even when the player actually makes a major error. This is intrinsic to supervised learning on human games: blunders are low-frequency events that receive little weight in the loss function. A model predicting the most common human move will always underrepresent the distribution's tail. This is a known limitation shared by all prior human chess models.

The slight accuracy decline in the $\geq$2000 bracket relative to 1900--1999 (57.38\% $\to$ 56.80\%) reflects the heterogeneity of the $\geq$2000 cohort, which spans club players to titled players with diverse opening preparation and stylistic preferences.

\section*{Acknowledgments}

The authors sincerely thank Microsoft for providing Azure compute credits and access to GPU resources. This support enabled the training of the models, extensive experimentation, and the development of the research presented in this paper.

The authors also express their gratitude to Lichess for generously making large-scale chess data publicly available. Their commitment to open data has been invaluable in enabling this research.

\clearpage \bibliography{references}

\end{document}